\documentclass{article}

\usepackage[english]{babel}
\usepackage{float}
\usepackage{graphicx}
\usepackage[letterpaper,top=2cm,bottom=2cm,left=3cm,right=3cm,marginparwidth=1.75cm]{geometry}

\usepackage{amsmath}
\usepackage{subfigure}
\usepackage[colorlinks=true, allcolors=blue]{hyperref}
\usepackage{indentfirst}
\usepackage{cite}

\makeatletter
\newif\if@restonecol
\makeatother

\usepackage[linesnumbered,ruled,vlined]{algorithm2e}%[ruled,vlined]{
\usepackage{algpseudocode}
\usepackage{amsmath}
\usepackage{authblk}
\usepackage{CJKutf8}
\title{Deep	Hierarchy Quantization Compression algorithm based on Dynamic Sampling}
\author[a]{Wan Jiang}
\author[a]{Gang Liu \thanks{Corresponding author: gliu@szu.edu.cn}}
\author[a]{Xiaofeng Chen}
\author[b]{Yipeng Zhou}

\affil[a]{College of Computer Science and Software Engineering, ShenZhen University}
\affil[b]{Department of Computing, Macquarie University}

\date{} % 去掉日期
\begin{document}
\begin{CJK}{UTF8}{gbsn}
\maketitle

\begin{abstract}
Unlike traditional distributed machine learning, federated learning stores data locally for training and then aggregates the models on the server, which solves the data security problem that may arise in traditional distributed machine learning. However, during the training process, the transmission of model parameters can impose a significant load on the network bandwidth. It has been pointed out that the vast majority of model parameters are redundant during model parameter transmission. In this paper, we explore the data distribution law of selected partial model parameters on this basis, and propose a deep hierarchical quantization  compression algorithm, which further compresses the model and reduces the network load brought by data transmission through the hierarchical quantization of model parameters. And we adopt a dynamic sampling strategy for the selection of clients to accelerate the convergence of the model. Experimental results on different public datasets demonstrate the effectiveness of our algorithm.
\end{abstract}

\section{Introduction}

\paragraph{}With the rapid growth of the number of intelligent devices worldwide and the progress of science and technology, the computing power and storage capacity of devices are also constantly improving, which allows smart devices to collect data at an unprecedented scale and speed. At the same time, deep learning also benefits from the improved ability of hardware devices to collect data, and has made great breakthroughs in the fields of image recognition, speech recognition, natural language processing, etc. The rapid development of smart devices has completely changed the way devices extract information from data. With the continuous growth of the amount of data, the scale of deep learning is also growing. To some extent, the breakthrough of deep learning can be attributed to the increasingly large datasets. Using the data collected from smart devices to improve the deep learning model has great potential. However, the collected data by smart devices are users' private data. We must consider the data privacy when using these data. If these private data are concentrated on the server for training, it will bring huge data security problems, Therefore, we usually don’t upload these data to the server for centralized training.
\paragraph{}In order to solve the data security problems caused by private data, Google proposed model-based average federal learning\cite{1}, federal learning , also known as federal optimization, in 2016, which allows multiple parties to cooperate in training models without sharing and collecting data, which is very compatible with local data storage and processing in the IOT. In order to cooperatively learn and share the global model, federal learning always keep all the data on the local device, federal learning iteration requires the participating nodes (also known as clients) to download the global model from the server, the participating nodes train their own model according to the local data, update the model parameters to the server, and then let the server aggregate the updates of the client to improve the model performance. Compared with traditional distributed learning, federal learning ensures the security of user privacy data, because the data will not leave the local device in the whole process \cite{10}. Federated learning allows users to keep their privacy data for training locally without requiring any participants to upload their privacy data to the central server. The main process of federal learning is shown in Figure 1.
\begin{figure}
\centering
\includegraphics[width=\textwidth]{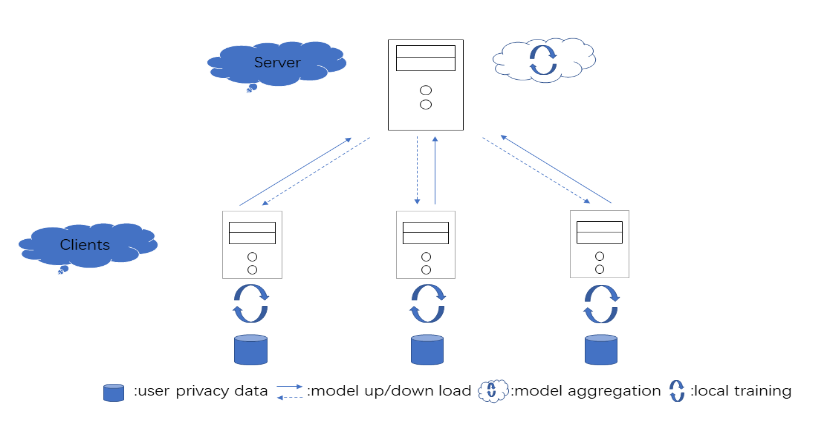}
\caption{\label{fig:frog}Federated learning. User data are stored securely on the user’s device to train a local model. Only the parameters from the local models are uploaded to the server for aggregation.}
\end{figure}
\paragraph{}The whole process can be simply divided into three steps: First, each participant downloads the latest aggregation model from the central server. Next, participants use their own local private data training model, which is the feature of Federated learning. The data is kept locally and does not need to be uploaded to the central server. Finally, the participants will upload the trained models to the central server, which will aggregate the models uploaded by each participant to form a new global model. These steps are repeated until certain convergence criteria are met. In the whole training process, the private data of participants will not leave the local, and the only communication with the central server is model update, which well protects the data privacy of users. However, in the process of communication, malicious users can still obtain users' data information through the information updated by the model\cite{2}. We can prevent privacy data from being stolen through some additional mechanisms, such as homomorphic encryption\cite{3}\cite{4}, or differential privacy\cite{5}.
\paragraph{}The training process of federated learning can be used for multiple machine learning models \cite{23} with the aim of minimizing the loss function they define. Common machine learning models are trained based on a large number of training samples, which consist of data and corresponding labels. The loss function on sample $j$ can be defined as $f(x_j,y_j,w)$, where  $x_j$, $y_j$ and $w$ represent the vector of input sample, sample corresponding label and current weight vector, respectively. For convenience, we use $f_j (w)$ to represent $f(x_j,y_j,w)$. Then, on the dataset $D_i$ of client $i$, the loss function can be defined as
\begin{equation}
    F_i (w) = \dfrac{\sum_{\substack{j \in D_i}} f_i(W)}{|D_i |}
\end{equation}

Stochastic gradient descent $(minibatchSGD)$ is used for training locally on the client. We use $W_i (t-1)$ represents the model of client $i$ in the $(t-1)$ round, then the update process of the model $W_i (t)$ in the $t$ round above the client can be expressed as follows.

\begin{equation}
    W_i(t) = W_i(t-1) - \alpha \bigtriangledown F_i(W_i(t-1))
\end{equation}
 $\alpha$ represents the learning rate during the training process, the model is uploaded to the server after training on the client, the server aggregates the client model to generate a new global model. The weight of the client needs to be considered during aggregation. The aggregation process can be expressed as follows, $W(t)$ is the final aggregation model.

\begin{equation}
    W(t) = \sum_{\substack{0<i<k+1}} \dfrac{|D_i|}{D_i}W_i(t)
\end{equation}

\paragraph{}An important challenge in federated learning is the communication cost in model transfer. Because during federated learning, every participating client must communicate a full model update during each training iteration. Each such update is the same size as the trained model, and its communication cost is non-negligible for modern architectures with millions of parameters. Since most of the federal learning devices in the actual scene are mobile phones, smart wearable devices, etc., the computing resources and network resources of these devices are limited. If the communication bandwidth is limited or the communication cost is too high, joint learning may become ineffective or even completely infeasible. In order to improve communication efficiency. At present, the work of reducing communication costs mainly focuses on two aspects: one is to reduce the amount of data carried in each round of communication, and the other is to reduce the number of communications between the client and the server. To reduce the amount of data in each round of communication, we quantify the model parameters in a deep hierarchy by analyzing the numerical distribution of the model parameters on the basis of the existing compression algorithm. This greatly reduces the amount of communication data in the gradient transmission process, but the compression of the model gradient through gradient quantization is a lossy compression method, which will undoubtedly bring the loss of model accuracy. So we discuss the second aspect of reducing communication costs. At present, the work to reduce the number of communication rounds is mainly through static sampling of the client\cite{6}, and static sampling selects a fixed proportion of clients for communication every time. In this paper, dynamic sampling is used to replace the original static sampling method, and the accuracy loss caused by lossy compression is compensated through the selection of a high proportion of clients in the early communication, and dynamic sampling accelerates the convergence of the model. Our contributions are as follows:
\begin{enumerate}
    \item Based on the existing communication compression algorithm, we deeply analyze the distribution of model parameters, and propose a deep Hierarchy quantization compression algorithm, which further reduces the communication cost.
    \item We use a dynamic sampling strategy of exponential annealing to replace the original static sampling strategy, which not only makes up for the adverse impact of deep level quantization compression on the model, but also accelerates the convergence speed of the model.
    \item Experiments on different public datasets verify the feasibility of our proposed algorithm.
\end{enumerate}
\section{Relate Work}
\subsection{Gradient Sparsification}
\paragraph{}Strom\cite{7} proposes to send only gradient updates larger than a predefined constant threshold, however the selection of the threshold is not easy and the threshold changes over time during training. Therefore Dryden et al.\cite{8} propose to select a fixed proportion of positive and negative gradients for updating. Aji and Heafield\cite{9} proposed a heuristic gradient sparse algorithm. The algorithm selects a threshold, which usually takes the kth largest gradient. During the communication process, the client only uploads gradients that are greater than the threshold, and gradients that are less than the threshold are retained locally for the next upload. Gradients stay local for the next communication, which greatly reduces the communication cost during training. In order to alleviate the impact of the method proposed by Aji and Heafield on the model performance, Lin and Han et al. proposed a deep gradient compression algorithm\cite{10}. Deep gradient compression uses momentum correction, pre-training, etc. to reduce the impact of gradient sparsification on model performance. The experimental results of the deep gradient compression algorithm in CNN image classification (CIFAR10\cite{17} and ImageNet\cite{19}), RNN language model (Penn Treebank \cite{20}), speech recognition (Librispeech Corpus \cite{21}) show that the lossless compression can be up to 600x.
\subsection{Gradient Quantization}
\paragraph{}Quantizing the gradient to a low precision value can reduce the communication bandwidth. Seide et al.\cite{11} proposed 1-bit SGD to reduce the size of gradient transmission data, and achieved 10x acceleration in traditional voice applications. Alistarh et al.\cite{12} proposed another method called QSGD, which makes a trade-off between accuracy and gradient accuracy. Similar to QSGD, Wen\cite{13} developed terngrad using a three-level gradient. These two works have proved the convergence of quantitative training, although terngrad only detects CNN and QSGD only detects the training loss of RNN. Others have tried to quantify the entire model, including gradients. The algorithm proposed by Zhou et al.\cite{14} uses 1-bit weight and 2-bit gradient.
\section{Our Work}
\paragraph{}In this section, we will introduce the specific process of the proposed algorithm in several modules.
\subsection{Gradient Sparsification}
\paragraph{}In federal learning, the client needs to upload the model to the server after training it locally. Before uploading the model, we perform threshold sparsification on the model to filter out the gradients to be uploaded. Specifically: before uploading the local model to the server, we perform a one-dimensional vectorization expansion of the model to be uploaded. Then we use the preset sparsification factor $k$ to select the top $k$$\%$ of the gradient of the whole model as the gradient of the model to be uploaded, and the absolute value of the $kth$$\%$ of the gradient value as the threshold $thr$. the remaining part of the gradient of the unuploaded model we do not discard directly, but keep it in the local client for gradient accumulation, waiting for the next communication. We use $Mask$ matrix for the sparsification operation, and the specific value of $Mask$ is taken as shown in Equation 4. Where $abs(W[i])$ represents taking the absolute value of the gradient.
\begin{equation}
    Mask[i]=\left  \{
    \begin{array}{ll}%ll按顺序是公式左对齐和条件左对齐
    1 & \text{$abs(W[i]) \ge thr$}\\
    0 & \text{$abs(W[i]) < thr$}                         
    \end{array}
\right.
\end{equation}
\paragraph{}We give an example to illustrate the sparsification process,as shown in Figure 2. The $model$ represents the original model, and the $Mask$ matrix is generated from the selected threshold gradient based on the position information of the original model. $Model$ represents the sparse model.
\begin{figure}[H]
\centering
\includegraphics[width=\textwidth]{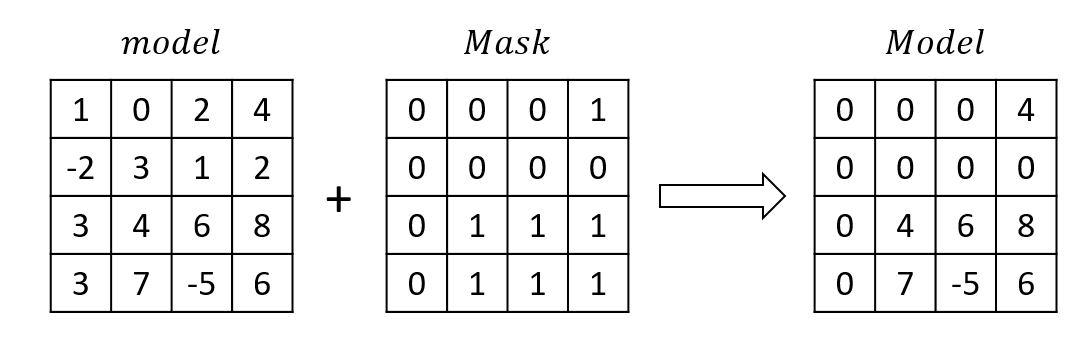}
\caption{Model gradient sparsification process. In this example, the threshold gradient is 4.}
\end{figure}
\subsection{Hierarchy Quantization Compression}
\paragraph{}The original model is sparsified to reduce the communication cost, but still did not reach the target. To further reduce the communication cost during model transmission, we quantify the sparsified model. First, we discuss the traditional quantization strategy. The quantization of traditional federated learning uses a small number of bits to quantize the model parameters. Although this quantization strategy greatly reduces the communication cost, it result in a loss of model performance. Our goal is to get a balance between model performance and model compression rate. By studying the distribution pattern of the gradients to be transmitted, we adopt a multi-bit encoding approach in order to ensure that the quantized model gradients do not deviate too much from the original gradients during quantization. This strategy does not perform as well as traditional low-bit quantization in terms of a single quantization, but combined with our previous model sparsification operation, so the communication cost from using multi-bit quantization coding will not exceed that of other low-bit quantization compression algorithms. The specific process of quantification can be described as follows: we find the gradient $\partial$ with the largest absolute value from the gradient after vectorization and set it as the upper limit of the gradient. Then we divide the gradient in the range of $thr$ and $\partial$ intervals, and the gradient interval we divide is encoded using 4 bits, where the first bit of the encoding represents $sign$ and the remaining three bits represent the interval number $num$. Next, we quantize the gradient in the corresponding interval range as the combination of the interval number and $thr$ corresponding to the gradient sign. Equation 5 describes the detailed quantification process, where $W[i]$ denotes the gradient to be quantized.$location$ represents the offset position of $W[i]$ compared to $thr$, and $sign$ represents the gradient sign.

\begin{equation}
    W[i]=\left  \{
    \begin{array}{ll}%ll按顺序是公式左对齐和条件左对齐
    thr+sign\times location\times (\dfrac{\partial-thr}{num})& \text{$W[i] \ge 0$}\\ 
    \\
    sign\times abs(thr+location\times (\dfrac{\partial-thr}{num}))& \text{$W[i] < 0$}   
    \end{array}
\right.
\end{equation}
 \paragraph{}The server aggregates the model gradients uploaded by all participating clients to form a new global model, and distributes it to all clients to start the next round of training until the model converges.

\renewcommand{\algorithmicrequire}{\textbf{Input:}}  % Use Input in the format of Algorithm
\renewcommand{\algorithmicensure}{\textbf{Output:}} % Use Output in the format of Algorithm
\begin{algorithm}
    \caption{Deep Hierarchy Quantization Compression Algorithm(DHQC)}
    \LinesNumbered
    \KwIn{Trained model: $W={W[0], W[1],.., W[M]}$, Sparsity factor: $k$, Currently selected clients:$C={1,2,…,N}$, Hierarchical quantization compression: $HQC$.}
    \KwOut{Local models waiting to be uploaded.}
    \For{$i=0;i \le M;i++$}{
         $thr \leftarrow k\% \; of \; W$\\
         $Mask[i] \leftarrow [W[i]] > thr$\\
         $\widetilde{W[i]} = W[i] \bigodot Mask[i]$\\
         $W[i] = W[i] \bigodot \neg Mask[i]$\\
         \For{$v \; in \; \widetilde{W[i]}$}{
            $max = max(abs(v),max)$\\
         }
         $end \; for$\\
         \For{$v \; in \; \widetilde{W[i]}$}{
            $v \leftarrow HQC(v)$\\
         }
         $end \; for$\\
    }
    $end \; for$\\   
    $Compressed \; model: \;\widetilde W \leftarrow \sum_{i=0}^{M}encode(w[i],Mask[i])$\\
    $Send \; compressed \; model:\;\widetilde W  \; to \; server$
\end{algorithm}

\paragraph{}Based on the gradient sparsification and multi-bit quantization compression strategy, we propose the Deep Hierarchy Gradient Compression algorithm (DHQC). Algorithm 1 gives the pseudo-code of the proposed algorithm flow, where $\odot$ denotes Hadamard Product, and the specific role of Mask[i] matrix is to mask the gradient to be uploaded for sparsification. ¬Encode means to encode the gradient.

\subsection{Dynamic Sampling}
\paragraph{}Mask sparsification and multi-bit quantization strategies can compress the communication cost even further, but this lossy compression still has an impact on the model performance, so we compensate for the accuracy loss caused by lossy compression by improving the sampling method. Static sampling \cite{6} is a common sampling strategy in federation learning. Static sampling uses a fixed sampling rate to select clients throughout the model training process, aggregates a new model by a fixed proportion of client models, and then distributes the new model to all clients, which is easy to implement but is not conducive to fast model convergence. In order to accelerate the convergence of the model, we adopt a dynamic sampling strategy. Specifically, we adopt a high sampling rate at the beginning of training, and gradually reduce the sampling rate as the number of communications increases. We allow as many client models as possible to participate in the model aggregation at the beginning of training, and obtain a more generalized model by involving a high proportion of clients in the initial training, and then dynamically reduce the number of participating clients to accelerate the convergence of the model. Although our dynamic sampling strategy has a large communication cost in the pre-training period, the number of selected clients decreases rapidly after a few rounds of communication. Choosing different decay rates according to the actual situation can control the communication cost of dynamic sampling not to exceed the communication cost of static sampling. We control the model sampling rate by an exponential function, as shown in Equation 6.
\begin{equation}
    R(\varphi,t) = \dfrac{M}{EXP(\varphi t)}
\end{equation}
$M$ represents the total number of clients, $R$ represents the proportion of each round of communication clients participating in training, $t$ represents the communication round, and $\varphi$ represents the attenuation rate. With the increase of communication times, the number of client samples decreases gradually. After several rounds of communication, the low sampling rate will cause the number of clients we select each time to be less than 1, which is not in line with our expectations. Therefore, we will limit the number of clients we select to at least 5 clients, so that our model can be aggregated normally.

\begin{figure*}[htbp]
    \subfigure[result on mnist]{
    \begin{minipage}[b]{0.45\linewidth}
        \centering
        \includegraphics[scale=0.45]{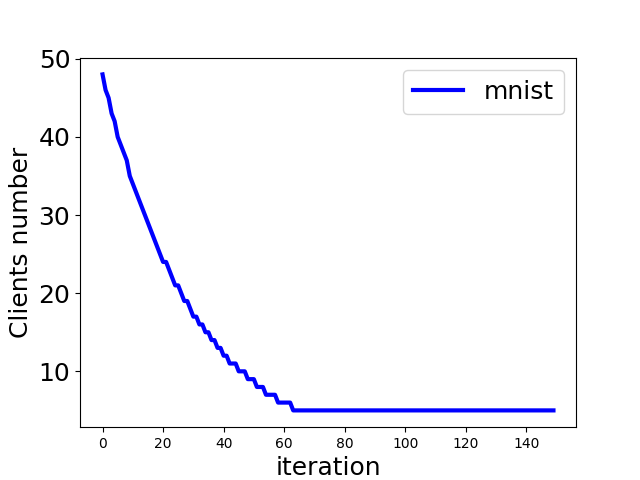}
    \end{minipage}
    }
    \hspace{30pt}
    \subfigure[result on cifar10]{
    \begin{minipage}[b]{0.45\linewidth}
        \centering
        \includegraphics[scale=0.45]{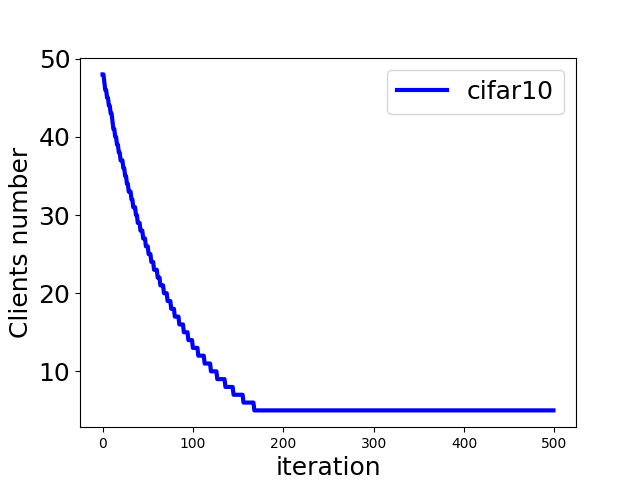}
    \end{minipage}
    }
    \caption{Comparison of client sampling rate and number of communications on the mnist dataset and the cifar10 dataset.}
\end{figure*}

\section{Experiments}
\paragraph{}In this section we will evaluate the performance of the proposed method on a widely used benchmark datasets. We have established an experimental comparison with the standard federated learning algorithm and other classical communication compression algorithms in image classification tasks. The performance of the model is evaluated by comparing the accuracy of the model with the theoretical compression ratio of the model. In addition, we also pay attention to the convergence speed of the model in the training stage. Next, we will give the experimental setup and experimental results.
\subsection{Experimental Setup}
\paragraph{}To explore the performance of the proposed algorithm in this paper, this section conducts comparative experiments using two classical datasets in the image classification domain, and to better represent the performance of our algorithm, we experimentally compare the proposed algorithm with other efficient communication algorithms in federation learning. Federated learning involves communication between distributed clients and a central server, but due to computational resource constraints, we only use the federation learning setup on a single Linux server to simulate real scenarios, ignoring communication noise and latency in the network. We give details of the experimental configuration. 
\subsubsection{Compare algorithms}

\paragraph{}The algorithms used in the experiments as follows:\\
\indent Baseline\cite{15}: A typical federated learning algorithm.\\
\indent DGC\cite{10}: A typical gradient sparse compression algorithm. The main idea is to reduce the communication cost by uploading only part of the gradient that is greater than the threshold.\\
\indent TernGrad\cite{13}: A typical gradient quantization algorithm. The main idea is to reduce the communication cost by quantifying the gradient into three values {-u, 0, u} (u is the maximum value of the gradient).\\
\indent DHQC: Our algorithm.

\subsubsection{Datasets}
\paragraph{}We selected two representative benchmark datasets widely used in image classification:\\
\indent MNIST\cite{16}: it contains 60000 training and 10000 test gray handwritten image samples, a total of 10 classes, and the dimension of each image is 28 × 28. Because MNIST features are easy to extract, this datasets is mainly used to train small networks.\\
\indent CIFAR10\cite{17}: it contains 60000 color images of 10 objects from frogs to aircraft, 50000 for training and 10000 for testing. It is a widely used benchmark datasets.

\subsubsection{Model}
\paragraph{}In order to evaluate the performance of the above algorithms, we choose ALL\_CNN model in the experiments, ALL\_CNN is a new network model based on convolution network proposed by Jost Tobias springenberg et al. \cite{18}in 2015. Unlike previous convolution networks, ALL\_CNN completely abandoned the pooling layer and full connection layer in the previous CNN network, and replaced the pooling layer with a larger step convolution layer and the full connection layer with a convolution kernel of 1. The whole network is composed of convolution layers, that is why it is called ALL\_CNN. 
\subsection{Experiments Results}
\paragraph{}Federated learning involves communication between distributed clients and central servers. However, due to the limitation of computing resources, we ignore the communication noise and delay in the network, and only use federated learning settings on a single Linux server to simulate the real scene. For the image classification task, we use the ALL\_CNN model as the client model. We used CIFAR10 and MNIST to experiment on the ALL\_CNN model, In order to illustrate the effectiveness of dynamic sampling on our proposed algorithm. Firstly,  we compare the performance of the Deep Hierarchy Quantization Compression algorithm that using static sampling with other algorithms. Next  we only change the sampling strategy, and we switch from the static sampling strategy to the dynamic sampling strategy, keeping the datasets and model unchanged. In order to distinguish compression algorithms based on static sampling from compression algorithms based on dynamic sampling, we name the algorithm using static sampling as DHQC,and the compression algorithm using dynamic sampling as D-DHQC. We used convergence rate, compression ratio, and accuracy to evaluate the performance of our experiment.
\subsubsection{Static Sampling}

\paragraph{}We explore the performance of the Deep Hierarchical Quantization Compression algorithm (DHQC) using a static sampling approach on the MNIST dataset and the CIFAR10 dataset, and the model used for the clients in the experiments is the ALL\_CNN model. To simulate federal learning in a real environment, we select a fixed percentage of clients to participate in training for each communication. When dividing the data, the data we divide to each client is not repeatedly divided, and we assume that this part of data is private and each client owns a copy of it, and we ensure that the overall data distribution satisfies independent homogeneous distribution. We experimented with several communication efficient algorithms, setting the maximum number of communications to 150 on the MNIST dataset and 500 on the CIFAR10 dataset. The experimental results are shown in Figure 4.

\begin{figure*}[htbp]
    \subfigure[result on cifar10]{
    \begin{minipage}[b]{0.45\linewidth}
        \centering
        \includegraphics[width=7cm,height=4.5cm]{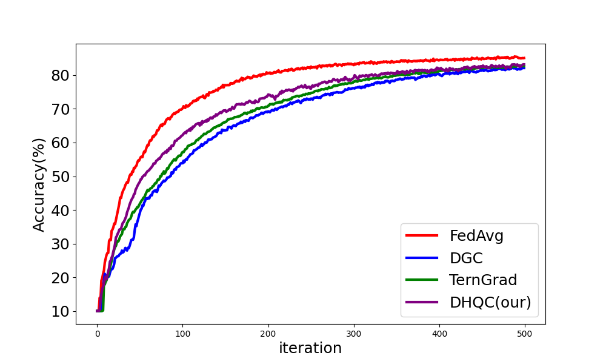}
    \end{minipage}
    }
    \hspace{20pt}
    \subfigure[result on mnist]{
    \begin{minipage}[b]{0.45\linewidth}
        \centering
        \includegraphics[width=7cm,height=4.5cm]{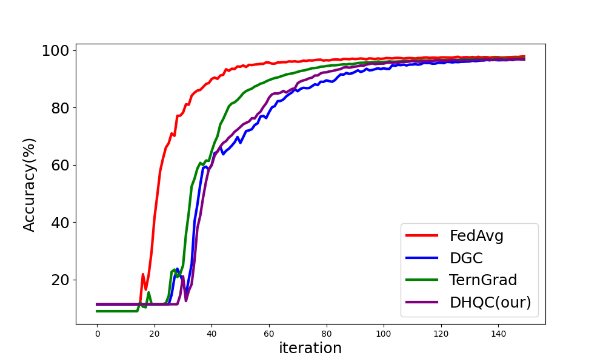}
    \end{minipage}
    }
    \caption{The accuracy comparison of communication rounds,based on (a) cifar10 dataset and (b) mnist dataset.}
\end{figure*}

\paragraph{}The experimental results of the model on two datasets when trained with different communication efficient joint learning methods are presented in Figure 4. Where (a) represents the experimental results of the model on the CIFAR10 dataset and (b) represents the experimental results on the MNIST dataset, with the x-axis representing the number of iterations and the y-axis representing the accuracy of the model. We observe that all communication compression methods achieve a fast convergence, which is very close to the uncompressed baseline (FedAvg) in the later stages of training, although not better than the baseline in the early stages of training. By analyzing the experimental results we can make the inference that the model is more sensitive to the performance loss caused by lossy compression in the early stage of training. Because the model is in an undertrained state. After several rounds of training, the model becomes a generalized model, when the gradients of the model tend to decrease in the same direction, so that in the later stages of training, all compression algorithms can achieve similar performance to the baseline algorithm. Next, we analyze the results of different compression algorithms in detail.\\
\indent FedAvg:The typical federal learning algorithm proposed in \cite{15}, which we use as the baseline algorithm in this experiment. As can be seen in Figure 4, the experimental results on different datasets demonstrate that the uncompressed baseline algorithm performs best in the early stage of training, which we attribute to the strong sensitivity of the model parameters to lossy compression in the early stage, and the complete model update facilitates the model to generate a generalized model in the early stage, but reaching the later stage of training, due to the increasing self-learning capability of the model, various communication compression algorithms are able to achieve similar accuracy, so in terms of model performance, the baseline algorithm does not have a significant advantage in the whole training process. Moreover, the baseline algorithm transmits the complete model update during the communication process, so it has the highest communication cost.\\
\indent TernGrad:The typical gradient quantization algorithm proposed in \cite{13}. The main idea is to reduce the communication cost by quantizing the gradient to three values $\{-u, 0, u\}$ ($u$ is the maximum value of the gradient). Theoretically, TernGrad can reduce the amount of communication passed from the working nodes to the parameter server by at least $32/log2(3)\approx20.18$ times. In fact, TernGrad needs at least 2 bits to encode \{-1,0,1\}, so it can actually reduce the traffic by about 16 times. The reason for the poor performance of TernGrad compared to the baseline algorithm in the early stages of training is the effect of lossy compression on the model parameters. After multiple iterations of training, TernGrad can achieve similar accuracy levels of the baseline algorithm in the later stages of training and reduces the transmission cost in federal learning communication by quantization. However, the model compression rate of TernGrad is not satisfactory.\\
\indent DGC:DGC algorithms reduce communication costs by transmitting only important gradient updates, where they use the absolute magnitude of the gradient as an evaluation metric. Specifically: in each round of communication, DGC transmits only the gradient updates whose gradient changes lie in the top $k\%$, and smaller gradient updates are accumulated locally and await the next communication. From the experimental results, we can conclude that it is reasonable to judge the importance of the gradient based on the magnitude of the gradient change, and the performance of the model in the late training phase supports this conclusion. However, when we only transmits part of the model gradient, the remaining gradient gradient will lead to the loss of model information, which will lead to poor performance of the model in the early training period. As the experimental results on different datasets show, both sparsification and quantization of the model parameters have some adverse effects on the model convergence due to the sensitivity of the prior model parameters and the lossy compression of the model parameters.\\
\indent DHQC:Our algorithm. We further reduce the communication cost during communication by combine the sparse and multi bit quantization strategies. As the experimental results show, similar to other lossy compression algorithms, our proposed algorithm performs poorly in the early stage of training. However, with the training proceeds, the performance of the model achieves similar accuracy of the baseline algorithm in the late stage of training, which indicates that our proposed method is convergent.\\
\indent The performance of different algorithms on the model is given in the previous section. We analyze the experimental results and give the reasons for these experimental phenomena. Although our algorithms do not have a clear advantage in model performance, in the field of communication optimization we are more interested in another metric, which is the compression rate. In Figure 5 we give specific data on the compression ratio and model performance of the compression algorithm.

\begin{figure*}[htbp]
    \subfigure[model performance with different algorithms]{
    \begin{minipage}[b]{0.45\linewidth}
        \centering
        \includegraphics[width=7cm,height=4.5cm]{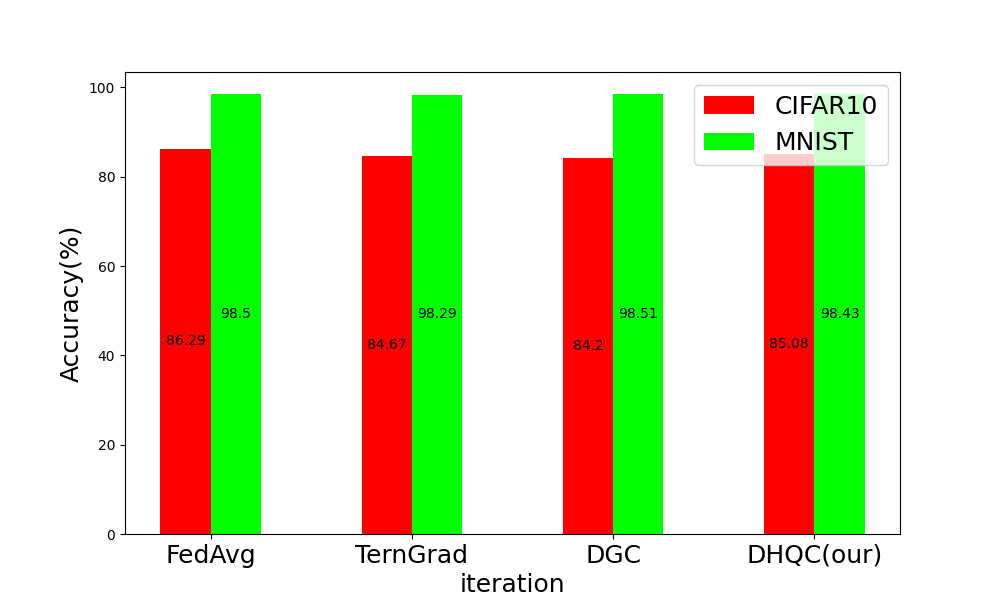}
    \end{minipage}
    }
    \hspace{20pt}
    \subfigure[compression ratio with different algorithms]{
    \begin{minipage}[b]{0.45\linewidth}
        \centering
        \includegraphics[width=7cm,height=4.5cm]{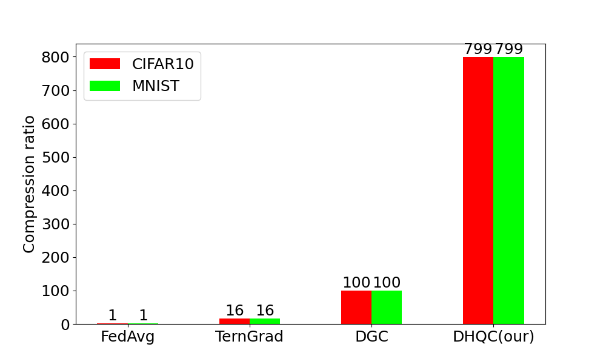}
    \end{minipage}
    }
    \caption{(a) represents the performance of different communication efficient algorithms on different image classification tasks. Where the x-axis represents the different algorithms and the y-axis represents the model performance. (b) represents the compression ratios that can be achieved by different algorithms during complete communication. Where the x-axis represents the different compression algorithms and the y-axis represents the compression ratio.}
\end{figure*}

\paragraph{}As shown in Figure 5, the model performance of our algorithm on the CIFAR10 dataset is reduced by 1.21\% compared to the baseline, but outperforms the other compression algorithms. The experimental results for all algorithms on the MNIST dataset are very close. Although we use more bits for encoding compared to other quantization compression algorithms, there is no increase in the cost of the communication process because we perform a sparsification operation on the model before quantizing it. Compared to the baseline, our algorithm can achieve a compression ratio of 799 times, and also far outperforms other compression algorithms.
From the experimental results on two image classification tasks we have the following conclusions.
\begin{enumerate}
    \item Our algorithm outperforms other communication efficient algorithms in terms of theoretical compression rate metrics with guaranteed model performance, which saves a significant amount of communication costs.
    \item The experimental data in Figs. 4 and 5 demonstrate that our algorithm achieves a satisfactory balance between compression rate and model performance.
\end{enumerate}

\subsection{Dynatic Sampling}

\paragraph{}We can infer from the previous experiments that more model information is more beneficial to the convergence of the model in the early stage of federal learning training. Therefore, we should transfer more model information in the early stage of training, which is beneficial to the convergence and training speed of the model in the early stage. Based on this conclusion, in order to reduce the impact of lossy compression on the model performance in the early stage and to speed up the convergence of the model, we improve the model performance by changing the client selection strategy. Specifically: we use a dynamic sampling strategy instead of the original static sampling. The reason for using the dynamic sampling strategy is that by selecting a high proportion of clients to participate in model aggregation in the early stage, the global model can obtain more information from a high proportion of clients in the early stage of training, which facilitates the model to generate a more generalized model in the early stage.
To ensure the validity and fairness of the experiments, we keep all settings unchanged except for changing the clients sampling method. We conduct experiments on the MNIST dataset and the CIFAR10 dataset for the depth hierarchical quantization compression algorithm using different sampling strategies. DHQC represents the deep	hierarchy quantization compression algorithm using static sampling, and D-DHQC represents the deep hierarchy quantization compression algorithm using dynamic sampling.

\begin{figure*}[htbp]
    \subfigure[result on cifar10]{
    \begin{minipage}[b]{0.45\linewidth}
        \centering
        \includegraphics[width=7cm,height=4.5cm]{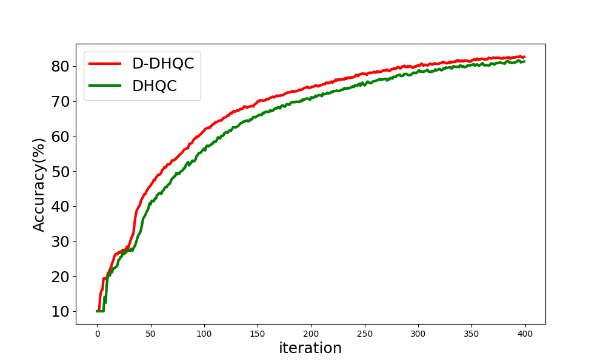}
    \end{minipage}
    }
    \hspace{20pt}
    \subfigure[result on mnist]{
    \begin{minipage}[b]{0.45\linewidth}
        \centering
        \includegraphics[width=7cm,height=4.5cm]{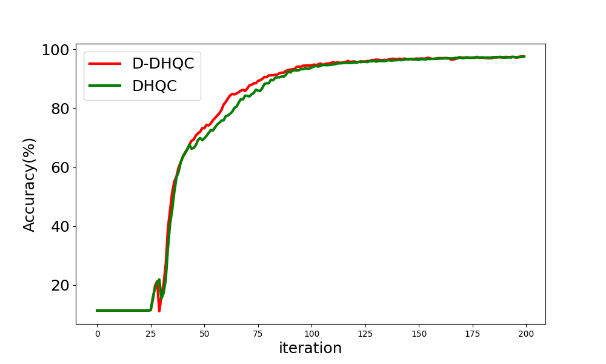}
    \end{minipage}
    }
    \caption{The accuracy comparison of communication rounds,based on (a) cifar10 dataset and (b) mnist dataset.}
\end{figure*}

\paragraph{}Figure 6 shows the experimental results of the deep hierarchy quantization compression with different sampling strategies on two image classification datasets, where the x-axis represents the number of communications and the y-axis represents the accuracy of the model on the test set. We set the maximum number of communications to 400 rounds on the CIFAR10 dataset and 200 rounds on the MNIST dataset, and from the experimental results we can see that:\\
\indent Convergence speed: After change the clients sampling method, the experimental results on the CIFAR10 dataset show that our algorithm has a certain improvement in convergence speed. And there is no significant improvement on the MNIST dataset.\\
\indent Model accuracy: The algorithm using dynamic sampling strategy on the CIFAR10 dataset outperforms the traditional static sampling approach.\\
\indent Communication cost: Although the dynamic sampling strategy selects a high percentage of clients to participate in aggregation in the early stage of training. However, as the training proceeds, the number of clients sampled in each round of communication will gradually decrease until a pre-set minimum value. In this process, we control the communication cost by the decay factor will not exceed the communication cost in the static sampling method.\\

\section{Discussion}
\paragraph{}In this paper, we compare our algorithm with other classical communication compression algorithms on image classification tasks. The experimental results verify that our algorithm achieves a competitive performance compared to other algorithms both in terms of compression multiplicity and accuracy, which provides new ideas and directions for reducing communication costs in the federation communication process. For the sake of analysis, we ignore the complexities in real networks. It is undeniable that there are still more factors to be considered for federation learning in real scene, which requires further simulation experiments. Modern deep neural models have a large number of parameters, such as VGG-16, whose total number of parameters exceeds 100 million. Due to the lack of computational resources, we can only perform simulation experiments on ALL\_CNN model, and our next work is to explore simulation experiments on large-scale deep neural networks with more computational resources.
\section{Conclusion}
\paragraph{}To achieve communication-efficient federation learning, this paper proposes a deep hierarchical quantized compression method to save communication costs. In order to reduce the impact of lossy compression on model performance, so we use dynamic sampling instead of static sampling in client selection. The improved algorithm using dynamic sampling has a good improvement in the prediction accuracy of the model compared with the static sampling algorithm, while not increasing the communication cost in the communication process. The results of the experiments conducted for the image classification task show that the proposed method achieves competitive results.

\section*{Acknowledgement}This work was supported by the Shenzhen Fundamental Research Program(20200814105901001)
\bibliographystyle{unsrt}
\bibliography{main}

\begin{thebibliography}{10}

\bibitem{1}
H.~B. Mcmahan, E.~Moore, D.~Ramage, and Bay Arcas.
\newblock Federated learning of deep networks using model averaging.
\newblock 2016.

\bibitem{10}
Y.~Lin, S.~Han, H.~Mao, Y.~Wang, and W.~J. Dally.
\newblock Deep gradient compression: Reducing the communication bandwidth for
  distributed training.
\newblock 2017.

\bibitem{2}
E.~Bagdasaryan, A.~Veit, Y.~Hua, D.~Estrin, and V.~Shmatikov.
\newblock How to backdoor federated learning, 2018.

\bibitem{3}
K.~Bonawitz, V.~Ivanov, B.~Kreuter, A.~Marcedone, and K.~Seth.
\newblock Practical secure aggregation for privacy-preserving machine learning.
\newblock In {\em the 2017 ACM SIGSAC Conference}, 2017.

\bibitem{4}
S.~Hardy, W.~Henecka, H.~Ivey-Law, R.~Nock, G.~Patrini, G.~Smith, and
  B.~Thorne.
\newblock Private federated learning on vertically partitioned data via entity
  resolution and additively homomorphic encryption.
\newblock 2017.

\bibitem{5}
Marti, N~Abadi, A.~Chu, Ian~J Goodfellow, Hugh~Brendan Mcmahan, I.~Mironov,
  K.~Talwar, and Z.~Li.
\newblock Deep learning with differential privacy.
\newblock {\em ACM}, 2016.

\bibitem{23}
E.~R. Ziegel and R.~Myers.
\newblock Classical and modern regression with applications.
\newblock {\em Technometrics}, 33(2):248, 1991.

\bibitem{6}
H.~B. Mcmahan, E.~Moore, D.~Ramage, S.~Hampson, and Bay Arcas.
\newblock Communication-efficient learning of deep networks from decentralized
  data.
\newblock 2016.

\bibitem{7}
N.~Strom.
\newblock Scalable distributed dnn training using commodity gpu cloud
  computing.
\newblock In {\em Interspeech 2015}, 2015.

\bibitem{8}
N.~Dryden, T.~Moon, S.~A. Jacobs, and B.~V. Essen.
\newblock Communication quantization for data-parallel training of deep neural
  networks.
\newblock In {\em 2016 2nd Workshop on Machine Learning in HPC Environments
  (MLHPC)}, 2016.

\bibitem{9}
A.~F. Aji and K.~Heafield.
\newblock Sparse communication for distributed gradient descent.
\newblock 2017.

\bibitem{17}
A.~Krizhevsky and G.~Hinton.
\newblock Learning multiple layers of features from tiny images.
\newblock {\em Handbook of Systemic Autoimmune Diseases}, 1(4), 2009.

\bibitem{19}
D.~Jia, D.~Wei, R.~Socher, L.~J. Li, L.~Kai, and F.~F. Li.
\newblock Imagenet: A large-scale hierarchical image database.
\newblock pages 248--255, 2009.

\bibitem{20}
M.~Maamouri, A.~Bies, T.~Buckwalter, and W.~Mekki.
\newblock The penn arabic treebank: Building a large-scale annotated arabic
  corpus.
\newblock 2004.

\bibitem{21}
V.~Panayotov, G.~Chen, D.~Povey, and S.~Khudanpur.
\newblock Librispeech: An asr corpus based on public domain audio books.
\newblock In {\em ICASSP 2015 - 2015 IEEE International Conference on
  Acoustics, Speech and Signal Processing (ICASSP)}, 2015.

\bibitem{11}
F.~Seide, H.~Fu, J.~Droppo, G.~Li, and D.~Yu.
\newblock 1-bit stochastic gradient descent and its application to
  data-parallel distributed training of speech dnns.
\newblock In {\em Conference of the International Speech Communication
  Association}, 2014.

\bibitem{12}
A.~Dan, J.~Li, R.~Tomioka, and M.~Vojnovic.
\newblock Qsgd: Randomized quantization for communication-optimal stochastic
  gradient descent.
\newblock 2016.

\bibitem{13}
W.~Wen, C.~Xu, F.~Yan, C.~Wu, Y.~Wang, Y.~Chen, and H.~Li.
\newblock Terngrad: Ternary gradients to reduce communication in distributed
  deep learning.
\newblock 2017.

\bibitem{14}
S.~Zhou, Y.~Wu, Z.~Ni, X.~Zhou, H.~Wen, and Y.~Zou.
\newblock Dorefa-net: Training low bitwidth convolutional neural networks with
  low bitwidth gradients.
\newblock 2016.

\bibitem{15}
H.~B. Mcmahan, E.~Moore, D.~Ramage, S.~Hampson, and Bay Arcas.
\newblock Communication-efficient learning of deep networks from decentralized
  data.
\newblock 2016.

\bibitem{16}
Y.~Lecun and L.~Bottou.
\newblock Gradient-based learning applied to document recognition.
\newblock {\em Proceedings of the IEEE}, 86(11):2278--2324, 1998.

\bibitem{18}
JT~Springenberg∗, A~Dosovitskiy∗, T.~Brox, and M.~Riedmiller.
\newblock Striving for simplicity: The all convolutional net.
\newblock {\em eprint arxiv}, 2014.

\end{thebibliography}

\end{CJK}
\end{document}